# Circle detection using isosceles triangles sampling

Hanqing Zhang[a], Krister Wiklund[a], Magnus Andersson [a,*]

[a]Department of Physics, Umeå University, Linneaus Vaeg 9, SE-901 87 Umeå, Sweden.

*Corresponding author: Magnus Andersson, magnus.andersson@umu.se, +46 90 786 6336

## Abstract

Detection of circular objects in digital images is an important problem in several vision applications. Circle detection using randomized sampling has been developed in recent years to reduce the computational intensity. Randomized sampling, however, is sensitive to noise that can lead to reduced accuracy and false-positive candidates. This paper presents a new circle detection method based upon randomized isosceles triangles sampling to improve the robustness of randomized circle detection in noisy conditions. It is shown that the geometrical property of isosceles triangles provide a robust criterion to find relevant edge pixels and thereby efficiently provide an estimation of the circle center and radii. The estimated results given by the isosceles triangles sampling from each connected component of edge map were analyzed using a simple clustering approach for efficiency. To further improve on the accuracy we applied a two-step refinement process using chords and linear error compensation with gradient information of the edge pixels. Extensive experiments using both synthetic and real images were presented and results were compared to leading state-of-the-art algorithms and showed that the proposed algorithm: are efficient in finding circles with a low number of iterations; has high rejection rate of false-positive circle candidates; and has high robustness against noise, making it adaptive and useful in many vision applications.

## 1. Introduction

Detection of circular objects is a fundamental feature extraction task in pattern recognition, which has continuously been developed to achieve better computational performance and accuracy [1], [2]. Several applications within the fields of; computer vision, physics, and biology, relies on detection of circular patterns, e.g., iris detection [3], cell counting [4], cell shape identification [5], and tethered particle tracking [6]. Apart from the execution time and accuracy, the major challenges when detecting circles are the presence of noise, low contrast, distortion and blurring boundaries, which are common in real images. These are either due to the settings of the image sensor or bad light conditions resulting in excessive amount of irrelevant textures, incomplete and distorted circle contours, and an added number of false-positives. Therefore, there is a need of circle detection methods with robustness against noise in images since these are handled rather poorly in conventional circle detection algorithms. The motivation and aim of this work is hence to improve the detection of circular shapes in the presence of noises.

The Hough transform (HT) was invented in the 50s and is a technique that can isolate features of a particular shape, e.g., circles in an image [7], [8]. In the circular Hough transform (CHT), an accumulator array is created for mapping the extracted edge pixels in the image to circle parameters in a 3-dimensional (Hough) space. This technique reaches a relatively high degree of accuracy but at the cost of the huge demand of memory space for the accumulator, and high computational complexity due to the time-consuming voting procedure. In addition, the indiscriminative voting procedure is also susceptible to noise for complex backgrounds.

Over decades, modifications of the CHT algorithm have been made to improve the accuracy and execution time. One strategy is to use geometrical constraints, where three commonly used constraints are; gradient-, chord-, and geometry-symmetry-based. The gradient-based approach [9] was proposed to accelerate the voting process. Ideally, the gradient of edge points on the circumference of a circle points at the center of the circle. This simple geometrical property can be used to reduce the dimension of a 3D accumulator to a 2D accumulator according to Yip *et al.* [10]; Rad *et al.* included both gradient and symmetric constraints for the selection of circles [11]; Jia *et al.* used a gradient based method to determine the circle center [12]; and Pan *et al.* calculated the gradient to build center projection lines in parameter space to locate the maximum intersecting points [13]. However, it is reported that using gradients is more sensitive to noise compared to using the edge points [14]. The chords-based methods find the centers at which perpendicular bisectors of the chords intersect. Better robustness against noise and distinctive probability distribution in the voting space can be found according to refs. [15, 16]. However, the computational complexities of calculating the intersecting point become a major drawback for such method. The global-symmetry of a circle can also be used to locate center points accurately and efficiently [17]. However, this property is based on the assumption that circles are ideal and complete, which is not very sufficient in handling occlusion and noisy conditions. Global symmetry are often used in combination with other properties to enhance the accuracy and efficiency [18].

Besides using the geometrical constraints, the enumerative voting process of the traditional Hough transform has been modified using randomized sampling for efficiency. The Randomized Hough transform (RHT) replaces the fundamental voting procedure for HT-based methods with random sampling of three edge points, and apply converging mapping as well as dynamic storage [19]. Based on a randomized sampling strategy several refined methods have been developed. Chen and Chung proposed a Randomized circle detection (RCD) algorithm using four edge points sampling [20]. According to their experimental results, the performance of RCD outperforms RHT in low to modest noise levels. By using four sampling points instead of three, the sampling strategy in RCD generates considerably less number of circle candidates in the validation process, which reduce a large amount of computational overhead. Inspired by the four-points sampling strategy, with a fourth point as validity check, multiple-evidence-based sampling strategy is proposed to further reduce the computational overhead in GRCD-R and GLRCD-R [21].

Improved circle detection work-flow, using classification of edge candidates or generation of line segments, has also been shown to reduce the computation time and efficiently find circle candidates. The EDCircles algorithm [2] achieves parameter-free, real-time performance using line segments generated by the Edge Drawing Parameter Free (EDPF) algorithm [22]. The edge segments were also applied in [23] along with a top-down scheme to enhance efficiency. Isophote curvature analysis was applied in [24] as a classification of edge pixels that improved the robustness and had limited dependency from the edge extractor. All these methods obtain decent performance and show certain robustness against noise.

In this paper, we present a randomized multi-circle detection algorithm based on a novel sampling process using geometrical constraints for searching valid circles. Our proposed method replaces the conventional three or four edge pixel strategies with identification of isosceles triangles (ITs) using gradient information of the edge points. The geometrical properties of ITs have several advantages. First of all, ITs can be detected from any pair of edge points on a circle, resulting in high probability for finding true circles. Secondly, the geometrical constrains of ITs are capable of suppressing the false-positives from background noise and unrelated textures. Thirdly, the calculation of the ITs constraints has low computational complexity for the sampling process. In support of the special property of ITs features, the algorithm first collects pixels in connected components formed by the rule of connectivity of edges pixels, then applies a sampling process with clustering in parameter space using ITs features. Based on the collected ITs, circle candidates are calculated and refined with chords and a linear error compensation with inliers verified using gradients. In the end, each candidate circles are verified with a sector based validation process. We apply our method using synthetic images as well as real images to demonstrate its efficiency and robustness in the presence of noise.

## 2. Background

We briefly introduce the fundamental sampling strategies used in many state-of-the-art randomized circle detection algorithms, where two commonly used strategies are the three-points circle sampling [17, 18] and the four-points circle sampling [19, 20, 24]. We start by introducing the basics sampling strategies and their properties.

### 2.1. Basic circle sampling strategy

The three-points circle sampling strategy randomly takes three edge pixels at each iteration to represent a circle. Suppose the coordinates of edges are $(x_1, y_1)$, $(x_2, y_2)$ and $(x_3, y_3)$. The center $a$ and $b$ and the radius $r$ can be obtained by [20],

$$a = \frac{\begin{vmatrix} x_2^2 + y_2^2 - (x_1^2 + y_1^2) & 2(y_2 - y_1) \\ x_3^2 + y_3^2 - (x_1^2 + y_1^2) & 2(y_3 - y_1) \end{vmatrix}}{4((x_2 - x_1)(y_3 - y_1) - (x_3 - x_1)(y_2 - y_1))} \tag{1}$$

$$b = \frac{\begin{vmatrix} 2(x_2 - x_1) & x_2^2 + y_2^2 - (x_1^2 + y_1^2) \\ 2(x_3 - x_1) & x_3^2 + y_3^2 - (x_1^2 + y_1^2) \end{vmatrix}}{4((x_2 - x_1)(y_3 - y_1) - (x_3 - x_1)(y_2 - y_1))} \tag{2}$$

$$r = \sqrt{(x_i - a)^2 + (y_i - b)^2}, \text{ for any } i = 1,2,3. \tag{3}$$

In practice, the calculation of $4((x_2 - x_1)(y_3 - y_1) - (x_3 - x_1)(y_2 - y_1))$ is evaluated before deriving $a$ and $b$ in order to avoid division by zero where three edge points are collinear. This is also the only way of rejecting false candidates in three-points sampling.

In four-points circle sampling, a fourth edge pixel $(x_4, y_4)$ is selected each iteration. This edge pixel is used to estimate a distance that determines whether it is located close enough to a circle. The distance is calculated as follows,

$$d = \left| \sqrt{(x_4 - a)^2 + (y_4 - b)^2} - r \right| \tag{4}$$

If the absolute value $d$ satisfies a threshold $T_r$ which defines the valid distance to form a circle, $d \leq T_r$ , a circle based on $a$, $b$ and $r$ can be chosen as a candidate. In practice, each edge points are separated with a minimum distance to avoid false positives under extreme cases during randomized sampling. The validation of this fourth points is reported to be able to discard 95% of bad sampled candidates, which lead to a huge boost in performance [21]. However, it is not able to validate true circles when the noise level is high and the edge map is filled with randomly distributed noises. These above mentioned sampling approaches are based on the assumption that a circle exists among sampled edge points without further checking of any other evidence. It is therefore susceptible to noise and can lead to false detection of circles.

## 2.2.    Solutions to problems

To overcome the above mentioned shortcomings, additional information of each edge point is necessary. Properties regarding the quality and validity of sampled edge points are often used for checking the edge point candidates. Such improvements include gradient information, symmetry of circles and chords. On one hand, these geometrical properties help identify true circles in a more robust and accurate way. On the other hand, each of these geometrical properties has its own weakness: the gradient estimation suffers from poor accuracy in the presence of noise; using symmetric properties restrict the number of valid edge points especially when distortion or occlusion occurs in an image and therefore not every point on a circle can contribute to finding the circle using the symmetric property; Chords-based methods have high probability distribution for circles, but the computational complexity restricts its performance. These problems lead to the need of finding a sampling strategy that has low computational complexity and high rejection rate against noise while maintaining a high probability of finding true circles. Our proposed ITs sampling satisfies all these properties.

## 3.   The proposed isosceles triangle sampling strategy

The general framework of the ITCiD algorithm is to identify isosceles triangles (ITs) inside circular objects using gradient information and specific constraints of ITs. Our sampling strategy, for identification of circle candidates, is divided into two stages. In the first stage, gradients in the x and y direction of the image are calculated. These vectors in tangent space are thereafter used to identify ITs in the second stage, which involves a detection process where two edge points are sampled from the edge map of the image and their corresponding geometrical properties to form an isosceles triangle are validated. The vector pairs that form an isosceles triangle are then used for estimation of center point and radius.

### 3.1.   Gradient orientation estimation

The gradient orientation is calculated using the intensity of the image. In an ideal situation without noise in the image, a vector that points towards the center of the circle candidate is achieved. The estimation of the gradient direction for each pixel in 2D plane is based on convolving the image with a Gaussian kernel as,

$$g_x(x, y) = D_x\big(G(x, y, \sigma)\big) * f(x, y) \qquad (5)$$

$$g_y(x, y) = D_y\big(G(x, y, \sigma)\big) * f(x, y) \qquad (6)$$

where $g_x(x,y)$ and $g_y(x,y)$ are the estimated gradient matrix in $x$ and $y$ direction, respectively. $D_x$ and $D_y$ are the gradient operators in the $x$- and $y$-directions, $G(x, y, \sigma)$ is the Gaussian kernel with standard deviation $\sigma$, and $f(x, y)$ is the intensity of the image. In practice, the gradient component $g_x(x,y)$ and $g_y(x,y)$ are normalized.

### 3.2.   Isosceles triangle detection

Isosceles triangles can be found by comparing the angle of triangle candidates using two gradient vectors. An example of an isosceles triangle formed from two gradient vectors at edge points A and B on the circle is shown in Fig. 1. The direction of the gradient vectors are evaluated using Eq. (5)

and (6). We apply a formula that relates the angles $\theta_1$ and $\theta_2$ with normalized gradient vectors $\overrightarrow{g(A)}$, $\overrightarrow{g(B)}$ and the vector $\overrightarrow{AB}$. This relation is given as,

$$\cos(\theta_1) = \frac{\overrightarrow{g(A)} \cdot \overrightarrow{AB}}{|\overrightarrow{AB}|} \qquad (7)$$

$$cos(\theta_2) = \frac{\overrightarrow{g(B)} \cdot \overrightarrow{BA}}{|\overrightarrow{BA}|} \qquad (8)$$

An isosceles triangle is found if $\theta_1$ and $\theta_2$ are similar. Instead of comparing $\theta_1$ and $\theta_2$ directly, the cosine of $\theta_1$ and $\theta_2$ are compared in order to reduce the computational time. In general, we assume that the gradient orientations of edge points on a single circle are all pointing toward or away from its center. This leads to the following criterion,

$$|cos(\theta_1) - cos(\theta_2)| \leq \Delta k \qquad (9)$$

where $\Delta k$ is the minimum absolute difference between values from Eqs. (7-8). If a pair of edge points fulfills the above condition an isosceles triangle is detected, otherwise the pair is discarded. The $\Delta k$ thus regulates the accuracy of triangles sampled and for perfect symmetric isosceles triangles, $\Delta k$ should be set to 0. However, due to the inaccuracy of the edge detection, gradient estimation, quantization error and noise, some degree of tolerance for error is needed in order to allow for more edge pairs in the sampling process. The value of $\Delta k$ should therefore be set according to the application since there is a trade-off between the ability to suppress false-positives detection and the capability to handle imperfect images.

In addition, the criterion can further be used to reduce the influence from undesired displacement of sampled edge points. For example, if two sampled edge pixels are too close to each other, the quantization errors of the gradient estimation reduce the accuracy in finding the true circle center. In this case, the corresponding vector pairs are parallel or close to parallel, thus unable to form a triangle. In addition, when the gradient vectors are parallel, edge pixels that satisfy Eq. (9) result in parallel lines. To avoid these false-positive detections, normalized vectors are tested with the following condition,

$$\left| \overrightarrow{g(A)} \cdot \overrightarrow{g(B)} \right| < \Delta p \tag{10}$$

where $\Delta p$ is a threshold value ranging from zero to one and refers to the minimum angels allowed between two vectors. The precision of angles are dependent on the kernel used in gradient estimation. Equation (10) is capable of preventing edge pixels that are in proximity to be selected, and also reject false positives from parallel lines and random noise. Equations (9-10) thereby set the criteria for finding valid ITs. It can be implemented by simply using three dot products of the vectors, three modulus operations, one division and one subtraction. In practice, the calculations are organized to be computational-efficient where only Eq. (7) is used for each edge pixel pairs while Eq. (10) is conditional, i.e., only used when Eq. (9) is fulfilled.

### 3.3.  Estimation of the center point and the radius

Based on the geometrical fact that any pair of points on a circle together with the circle center contribute to an isosceles triangle, it is easy to locate the center point of the circle using the geometrical property of the ITs, if the criteria in Eqs. (9-10) are fulfilled. In practice, our center estimation finds the intersecting point from the extension of normalized gradient vectors $\overrightarrow{g(A)}$ and $\overrightarrow{g(B)}$ based on the geometrical property of isosceles triangles. It requires only one pair of edge points in the calculation. This center estimation is fast and accurate for low noise situations. It is also obvious that the radius estimation can be conducted by using either the geometrical properties of ITs or Eq. (3) based on the estimation of center point.

## 4.   Proposed Algorithm and Implementation

In this section, an algorithm based on the ITs criteria for randomized circle detection as well as a sampling process for boosting the performance of randomized sampling is presented. The major steps include gradient calculation, edge classification, iterative ITs sampling, refinement and a validation process as shown schematically in Fig. 2. Detailed descriptions of each step are presented in the following subsections.

### 4.1.   Gradient Estimation and Edge Classification

The gradient of the image, in the x and y directions, is calculated according to Eq. (7) and Eq. (8), respectively. The strategy for designing the DoG kernel with proper standard deviation and size is a trade-off between the accuracy and speed of the algorithm. On one hand, the robustness of the gradient estimation against noise can be improved by applying a large kernel for Gaussian smoothing. On the other hand, small kernels ensure the computational efficiency of the algorithm.

In order to improve the sampling efficiency, instead of randomly sample edge pixels in the entire edge map, we apply a connected component approach where edge pixels are grouped into edge segments according to their connectivity with their neighbors. We adopt a 8-connected scheme, which implies that all pixels that touches the edges or corners of a pixel are considered as neighbors and all the neighboring pixels are chained to form an edge segment. The searching of edge pixels are thus limited to each local edge segment resulting in less iterations when sampling randomly.

### 4.2.   Sampling strategy

Our sampling strategy is based on a parameter space clustering process using ITs features. Edge pairs are randomly selected from a connected component and verified using the ITs criteria. To further increase the probability of finding circles the properties of all edge pairs that passed the ITs criteria are saved, i.e., the estimated center and radius, coordinates of the edge pairs, into a data structure. Thereafter, only the edge pairs that estimate a center point and radius within a given searching volume in parameter space are used to determine the edges that form a true circle. It is worth noting that the estimated centers and radii are distributed sparsely in parameter space, therefore, clustering of centers and radii is performed with the following routine. Assume we have an existing data structure with the estimated center, radius, and searching volume. If the center (x, y) and radius $r$ of a newly detected edge pair falls into the searching volume, the data structure will be updated by,

$$x_p = \frac{N * x'_p + x}{N + 1} \tag{11}$$

$$y_p = \frac{N * y'_p + y}{N + 1} \tag{12}$$

$$r_p = \frac{N * r'_p + r}{N + 1} \tag{13}$$

and

$$D_p = D'_p + \sqrt{(x_p - x'_p)^2 + (y_p - y'_p)^2 + (r_p - r'_p)^2} \tag{14}$$

where $N$ is the total number of edge pairs in the data structure, $x_p, y_p, r_p$ and $D_p$ represents the updated parameters of the center x, center y, radius, and searching range respectively while $x'_p$ , $y'_p$ , $r'_p$ and $D'_p$ are the values before update. If the properties of newly detected edge pair is not found in the searching volume of any existing data structures, a new data structure is created with a fixed searching range. There are three advantages of this process. First, this strategy helps preserve most of the valid edge pairs in the sampling process especially when there is limited number of edge points in a noisy image. For the purpose of our implementation using connected components, such strategy also reduces the misdetection especially when finding multiple circles in complicated edge segments. Second, the strategy is designed to reduce sampling iterations with efficient calculations. Instead of introducing complicated clustering methods such as k-means, region growing or voting in parameter space, which are computational expensive, the above binning approach solves most of cases regarding the sparse distribution of potential circle parameters estimated from a connected component. The update of the searching range ensures edge pairs that are forming the same circle to be efficiently collected within a limited number of iterations. Third, the update in the data structure also provides a certain degree of error tolerance against the imperfection of circles. Some edge points that deviate from an ideal circle can still be collected despite of noise.

### 4.3.    Gradient-dependent Refinement

The refinement process of our algorithm is based on the data structure constructed above and it is performed in two parts. The first part derives the center points using four edge points from two ITs sampled randomly in a data structure.  We demonstrate our scheme for locating the circle center with an example shown in Fig. 3, where two ITs, $A_1A_2C_A$ and $B_1B_2C_B$, are detected using the ITs criteria. However, imperfections and noises affect the shape of ITs and result in an estimation of the center that fall within a searching range, i.e., in the blue area. Instead of using the gradient estimation, we take the chord $A_1A_2$ and $B_1B_2$ and find the intersecting points O from their perpendicular bisectors. We denote the coordinate $A_1(x_1, y_1)$, $A_2(x_2, y_2)$, $B_1(x_3, y_3)$ and $B_2(x_4, y_4)$. We have,

$$a' = \frac{\begin{vmatrix} x_3y_3 - x_4y_4 & x_3 - x_4 \\ x_1y_1 - x_2y_2 & x_1 - x_2 \end{vmatrix}}{(y_3 - y_4)(x_1 - x_2) - (y_1 - y_2)(x_3 - x_4)} \tag{15}$$

$$b' = \frac{\begin{vmatrix} y_3 - y_4 & x_3y_3 - x_4y_4 \\ y_1 - y_2 & x_1y_1 - x_2y_2 \end{vmatrix}}{(y_3 - y_4)(x_1 - x_2) - (y_1 - y_2)(x_3 - x_4)} \tag{16}$$

where $a'$ and $b'$ are the center coordinate in the x and y direction, respectively. This scheme takes the advantage of high quality edge pair candidates using ITs sampling, and, at the same time, overcomes the weakness of estimation of the gradient direction. The probability for misdetection of the center coordinate using three points, due to the undesirable edge points positions as described in [20], is very low when using the above mentioned calculations with four points.

In the second part of the refinement, we use the linear error compensation method presented in [24], with the following modification. All inliers used in the error calculation are selected according to their gradient direction. Based on the circle specification from the first refinement stage, edge pixels in the connected component are selected within a certain distance $\Delta d$ from the circumference of a candidate circle, and the gradient direction of these edge pixels are compared. Since gradient estimation is sensitive to changes of the intensity in the image, this particular "alignment check" selects pixels that are affected less by errors from noise, low contrast and blurring boundaries. The inliers generated in this procedure are therefore more reliable.

## 4.4.    Validation process

After refinement a validation process is applied to minimize detection of false-positive candidates. The process divides a circle into several equal sized sectors, conducts voting from consecutive sectors, and checks the completeness of the circle. The center points of each sector on the potential circle circumference are collected and alignment of their gradients are controlled. Edge points and their corresponding sectors are marked valid if the condition for alignment is satisfied. We collect all of the valid sectors and validate the circle under the following principle. Consecutive sectors that are composed of at least three sectors are identified. Each consecutive sector contributes to a number of votes equal to the total number of sectors. We set restriction on the minimum number of votes with a pre-determined threshold value. The threshold controls the total amount of occlusions in a circle that the algorithm can tolerate. Candidate circles with voting numbers larger than the threshold are labeled as detected and all the edges that contribute to voting, during the validation process, are removed from edge segment. Otherwise, the sampling process restarts until it meet the ending condition.

## 5.   Experiments and Results

The evaluation of the proposed algorithm was conducted by experiments using both synthetic images and real images. Synthetic images were used for quantifying the properties of ITs sampling and real images were used for testing the general performance, accuracy and robustness against noise in comparison of three state-of-the-art approaches published in recent years. All internal parameters for the algorithms were fixed for all the experiments, i.e., the standard deviation of the DoG kernel and the kernel size was set to 1.28 and 5×5, respectively, whereas the parameters in the ITs criteria, were set to 0.1 and 0.92 for $\Delta k$ and $\Delta p$, respectively. We defined a completeness parameter to also exclude circles that had a high amount of occlusions and imperfections, i.e., >50 % of the pixels missing from a perfect circle. Finally, the maximum number of iterations was limited to the total number of edge pixels detected, using a Canny edge detector, in an image.

### 5.1.   Evaluation of IT sampling

In order to evaluate the properties of the proposed sampling method, an accumulator was created to quantify the probability of finding circles. We applied the sampling methods directly to the edge map of synthetic images and analyzed the distribution of votes generated in the accumulator, which thus estimates the circle center and radius in parameter space. For demonstrational purpose, we used a 2D accumulator to collect and quantify the voting of circle centers from sampling. In order to reduce the influence of uncertainty during the sampling process and present the probability of finding true circles, we performed 100 trials for each case and stored the average distribution of votes.

#### 5.1.1 Shape discrimination

First, we investigated the efficiency of discriminating circular objects in synthetic images containing various shapes, i.e., circular, triangular, ellipse, rectangular etc, without overlapping, as shown in Fig. 4. A circular shape with its center position at (30, 60) and with radius 20, can be seen on the left side of the image. In this experiment, instead of using connected components for efficiency, the entire edge map is used for deriving the distribution in a 2D accumulator array with the same size as the image. In Fig. 4B the distribution of possible circle centers using ITs sampling is shown. As can be seen, the distribution shows a significant peak at position (30, 60) indicating an estimation of the center position for the circle in Fig. 4A while considerably fewer votes are generated for the parallel lines, ellipse, squares, triangle, etc. Thus, the ITs sampling strategy can easily discriminate circular objects from other types of shapes.

### 5.1.2 Robustness using IT sampling

The property of robustness against noise is crucial for randomized sampling in many real-world applications. This property in ITs sampling was examined and compared with that of the conventional three points circle sampling and the most popular four-points circle sampling [23] and [24]. In order to demonstrate that the robustness against noise in the proposed algorithm is gained through the sampling method instead of a byproduct from parameter tuning, pre-processing, segments of connected-component or refinement and validation process, a series of experiments were conducted using a simple Sobel edge extraction, sampling in the edge map with a maximum iteration of 500 and the same smoothing factor for the distribution of votes collected in an accumulator. Synthetic images (256×256 pixels) of a single circle with different radius were created and Gaussian white noise were applied to the images using the MATLAB function 'imnoise' with zero mean and different levels of variance. The same edge map of image at each particular noise level was thereafter used by the three different sampling methods.

To quantify the results, we calculated the peak signal–to–noise (PSNR) ratio in the accumulator. The calculation of the PSNR in the accumulator follows the definitions of Gonzales [24]. The PSNR is calculated using the 2D accumulator in the absence of noises as a reference image $r(x, y)$ and the 2D distribution of the accumulators in the presence of noises as a test image $t_i(x, y)$, where i denotes the level of Gaussian noise. The size of the accumulator ($N_x$, $N_y$) was given the same size as the image. The PSNR is thus given as:

$$\text{PSNR(i)} = 10 \cdot \log_{10}\left[\frac{\sum_0^{N_x}\sum_0^{N_y}[\text{r}(x,y)]^2}{\sum_0^{N_x}\sum_0^{N_y}[\text{r}(x,y)-\text{t}_i(x,y)]^2}\right]$$

where the unit of PSNR is in dB. An example of PSNR results with a synthetic circle of radius 50 pixels are presented in Fig. 5. At low noise levels, the PSNR is high for all methods. Different parameters used in ITs sampling shows that the PSNR gets higher when $\Delta k$, i.e., the IT accuracy defined in Eq. (9), is set with a smaller value. As the noise increases, the PSNR of four-points sampling and three-points sampling decreases rapidly and signals are immersed into noise while ITs sampling is more robust to noise and provides relatively high PSNR until the noise level reach 0.2. The ITs sampling with a smaller $\Delta k$ tends to decrease faster as the amount of noises increases due to the limited sampling iterations. In our experiments, we found that in general ITs sampling provide better PSNR curves compared with four-points sampling, for circles with radius from 10 to 100 pixels. We denote the difference in PSNR between ITs sampling and four-points sampling at certain noise level as ΔPSNR. By using Δk=0.05 for ITs sampling, the maximum ΔPSNR at each circle radius is presented in Fig. 6. A positive ΔPSNR indicates that the PSNR in ITs sampling is better than the one with four-points sampling. As seen in the figure, for larger circle radius the ΔPSNR decreases. Even though the ITs sampling has certain scale dependency of circles, the overall results of ITs sampling is still better than that of the four-points sampling in the randomized circle detection.

## 5.2.  Evaluation of the ITCiD algorithm

The proposed ITCiD algorithm was applied to 10 test images found in the literature for comparison. The images consisted of; coins (255×256 pixels), ball (340×340), plates (400×300), cake (256×256), stability-ball (209×210), gobang (256×256), speaker (437×393), insulator (256×192), logo (283×344) and swatch (309×356). In Fig. 8 the results from the algorithm is shown. All experiments were conducted using an Intel CPU i7-4770 processor and the implementation of the algorithm was based on C++. Since many methods applied a preprocessing with undeclared or declared smoothing factor ranging from 1.0 to 1.5 [2], [23], [24], we also applied a smoothing factor of 1.0 using a 5×5 kernel.

### 5.2.1 Performance

To evaluate the general performance of the ITCiD algorithm, we first demonstrate the execution time as shown in Table.1. The execution times were evaluated by taking the average of 100 trails for each test image. In order to make a reasonable comparison with successful methods proposed recently, i.e. GRCD-R [21], EDCircles [2] and Isophotes [24], our execution times had been multiplied by a factor of 1.65 based on the benchmark comparison between our CPU and the one used in EDCircles. Comparing with GRCD-R, the results of the proposed algorithm show better computational performance, and more details in the image 'Logo' and 'Speaker' can be detected. The Isophotes is capable of finding circles efficiently using the isophotes curvature analysis and achieved boost in performance especially for images with few complex textures, e.g., 'Plate' and 'Stability-ball'. On the other hand, the speed of Isophotes is compromised when images contain complex textures, e.g., 'Coins' and 'Logo'. This was not the case for our proposed method. It can be observed from the table that ITCiD has a relatively stable execution time for these test images and the average execution is faster than Isophotes. The results also show that our method detected most circles within 30 ms. However, the performance of ITCiD is not as fast as EDCircles. The "bottle-neck" of our algorithm can be found in the analysis of the execution time for each part of our algorithm given in percentage, as presented in Table 2. The gradient calculation and contour extraction limited the overall performance and occupied almost half of the total execution time. The second most computational expensive part of the algorithm was the refinement process, which contributed to 21 % of the execution time. The refinement process helps determine circles precisely and cannot be removed from the algorithm. The consequence of lacking the refinement process in EDCircles is revealed in the accuracy evaluations. The sampling process, verification and other operations (labelled as 'Other' in Table 2) including memory assignments and update of contours took less than 30 % of the total execution time.

In order to quantitatively compare the efficiency of randomized sampling, we compared the results of possible circles $N_p$ (iterations) and the number of candidate circles $N_c$ (detected circles before refinement and validation) collected by RCD [20], GRCD-R [21], Isophotes [24] and our proposed ITCiD method. The result is shown in Table 3. It is clear that by applying classification of edge pixels before randomized sampling, the iterations have been greatly reduced for Isophotes and ITCiD. The classification using equal isophote curvature had better results in the reduction of total amount of possible circles. However, the 4 points sampling in Isophotes generate more false positives candidate circles, which limits its performance. With a slightly larger number of iterations, the number of circle candidates using ITCiD is less than achieved using Isophotes. By using ITs sampling, more false positive candidates were rejected and discarded and the total number of candidate circles was reduced. The reduced number of candidate circle improves the performance since less execution time will be spent on the refinement and validation process, making ITCiD an efficient algorithm.

### 5.2.2 Evaluation of the accuracy

The accuracy of the algorithms was measured following the convention from previously propose methods [24]. The ground truths of the circles in the 10 test images were measured using the circle Hough transform (CHT). The optimal parameters for CHT were set based on the pre-knowledge of radius sizes in each of the image. In Table 4, the averaged differences between parameters generated from CHT and those from GRCD-R, EDCircles, Isophotes and ITCiD are presented. The parameters include center coordinates ($\Delta a$, $\Delta b$) as well as the radius $\Delta r$. The results demonstrate subpixel accuracy for all the methods. It also shows that all the results generated from refinement strategies are more accurate than that generated through the verification process using arcs in EDCircle. There are arguments regarding whether the CHT provides the optimal ground truth of circle detection [24]. Since the ground truth can depend on the pre-processing, pre-determined parameters of different version of the circular Hough transform, and the imperfections of circles in the image, it is difficult to know which method that provides the best ground truth for real images at subpixel level. Therefore we compared the results only with the implementation of CHT and it was obvious that refinement process is necessary to ensure the accuracy to be close to the optimal.

### 5.3.    Robustness analysis

In this test, we conduct experiments on a series of images to demonstrate the robustness of our method against noise. We compared the detection results using our proposed algorithm with EDCircles on the noisy images as demonstrated in Fig. 8 and the quantitative comparison is presented in Fig. 9. When the Gaussian noise level is set to zero, the EDCircle can detect 19 circles and our method detects 11 circles due to the different implementation of minimum radius and circle completeness in our algorithm. As the noise level increase, the number of detected circles starts to decrease rapidly for EDCircles while our results show a much slower decrease. When the noise level reach $\sigma$=0.1, our method can still detect circles with a small radius while the EDCircles cannot detect any circles. Without any further help of denoising or rescaling the image, the detection of circles using EDCircles and ITCiD are limited by the edge extraction from either the EDPF algorithm used in EDCircles and the Canny edge detector used in our method. Even though the noise affects the edge detection, the ITs sampling strategy in the propose algorithm is capable of filtering out distorted or unrelated texture pixels, providing reasonable edge pairs within a limited number of sampling iterations and thus increase the probability of finding true circles. In addition, clustering of edge points as done using Eqs. (11)-(14) also provides high error tolerance and improves the efficiency.

We also compared our method with the Isophotes method for noise analysis and applied different levels of Gaussian noise, ranging from 0.01 to 0.30. Circle detection, under different amount of noise, are shown in Fig. 10, and the analysis is presented in Fig. 11. As can be seen, the overall number of detected circles using ITCiD is better than what is found using the Isophotes method. With gaussain noise level in regions [0.01, 0.10], the two plates with less contrast in the image are affected. It is clearly seen that ITCiD is more robust in the presence of noise than Isophotes within this noise level. When the noise level is between [0.10, 0.20], the edge extraction from the two high contrast plates are also affected. The ITCiD can still detect more circles due to the efficient sampling and clustering strategy. When the noise level reached up to 0.30 and more, the edge extraction is heavily affected and the connected component of edge segments used in our algorithm start to break apart, making it difficult for the sampling method to find true circles using small edge segment. On the other hand, the Isophotes based classification can group more pixels that belong to a circle but with higher false detection rate. In the Isophotes method, the number of false-positive candidates starts to rapidly increase for noise variance >0.16, and at ~0.30 more than 1 false-positive candidate is found. In contrast, the ITCiD algorithm has a low amount of false-positive candidates, <0.07 even though the noise level reaches 0.30.

## 6.   Conclusions

This paper presents a novel randomized circle detection algorithm based upon ITs sampling using the gradient of sampled edge pixels, combined with corresponding strategies using clustering and a two-step refinement process to achieve both efficiency and accuracy. The ITs criteria provides discriminative features for detection of circles and the particular sampling strategy require only a low number of iterations, has a high rejection rate of false-positive circle candidates, and is robust against Gaussian noise.

Detection of circles in real images were compared to the recent state-of-the-art algorithms GRCD-R, EDCircles and Isophotes, and the results show that the ITCiD achieves better performance with subpixel accuracy, in comparison to the randomized circle detection methods, and show in general better robustness against Gaussian noise. The proposed method can easily be combined with modern edge detection and contour extraction techniques to enhance the performance and accuracy of the detection of circles in vision systems such as robotic applications and for cell detection in biophysical experiments.

## 7.   Acknowledgements

We thank Dr. Ulrik Söderström for suggestions and comments on this work. This work was supported by the Swedish Research Council to M.A, 2013-5379.

## 9.  Figure captions

**Figure 1.** The figure shows how a pair of isosceles triangles are detected within a circle. Using two edge pixels A and B, and their corresponding gradient vectors (solid black arrows), which points at the center point, O, an isosceles triangle AOB can be identified.

**Figure 2.** A flowchart of the ITCiD algorithm.

**Figure 3.** The first step of center position refinement. Two edge pairs $A_1A_2$ and $B_1B_2$ fulfil the ITs criteria. $C_A$ and $C_B$ are the estimated center positions without refinement. The midpoint of the lines perpendicular to $A_1A_2$ and $B_1B_2$ however, intersect at point O, which provides a better estimate of the center of the circle.

**Figure 4.** A demonstration of a shape discrimination test. (A) Original image. (B) 2D voting distribution of center point of circle candidates using ITs sampling.

**Figure 5.** The PSNR vs. variance of the Gaussian noise in a 256×256 synthetic image containing a single circle with radius of 50 pixels. For low noise levels <0.01 the three methods perform equally well, whereas for noise levels >0.03, the ITs (blue) provides better PSNR than three (red) and four (green) point sampling.

**Figure. 6**. The maximum difference of PSNR between ITs sampling ($\Delta k$=0.05) and four-points sampling for given noise levels and circle radius.

**Figure 7.** Results of ITCiD using 10 real test images. The detected circles are marked in red with green dots at the centers.

**Figure 8**. Demonstration of ITCiD for different noise levels using a test image that contains circular shapes of different radius. The original image was exposed to different levels of Gaussian noise ranging from 0 to 0.1.

**Figure. 9**. Results of ITCiD using the image containing multiple circles as shown in Fig. 8. Since our method is based on randomized sampling, the results from the ITCiD are the average number of detections in 100 trails. The noise level on the x-axis was multiplied by 1000 for the comparison with EDCircles.

**Figure 10**. The ITCiD at different noise levels. The test image 'Plate' was used and exposed with different levels of Gaussian noise ranging from 0.01 to 0.3.

**Figure 11.** Noise tolerance experiment using the 'Plate' image. The number of correct and false circles detected in comparison to the ground truth for different noise levels. The red and purple lines are extracted data from [24] for comparison. The x-axis was multiplied by a factor of 100 to compare with the reference data used in Isophotes.

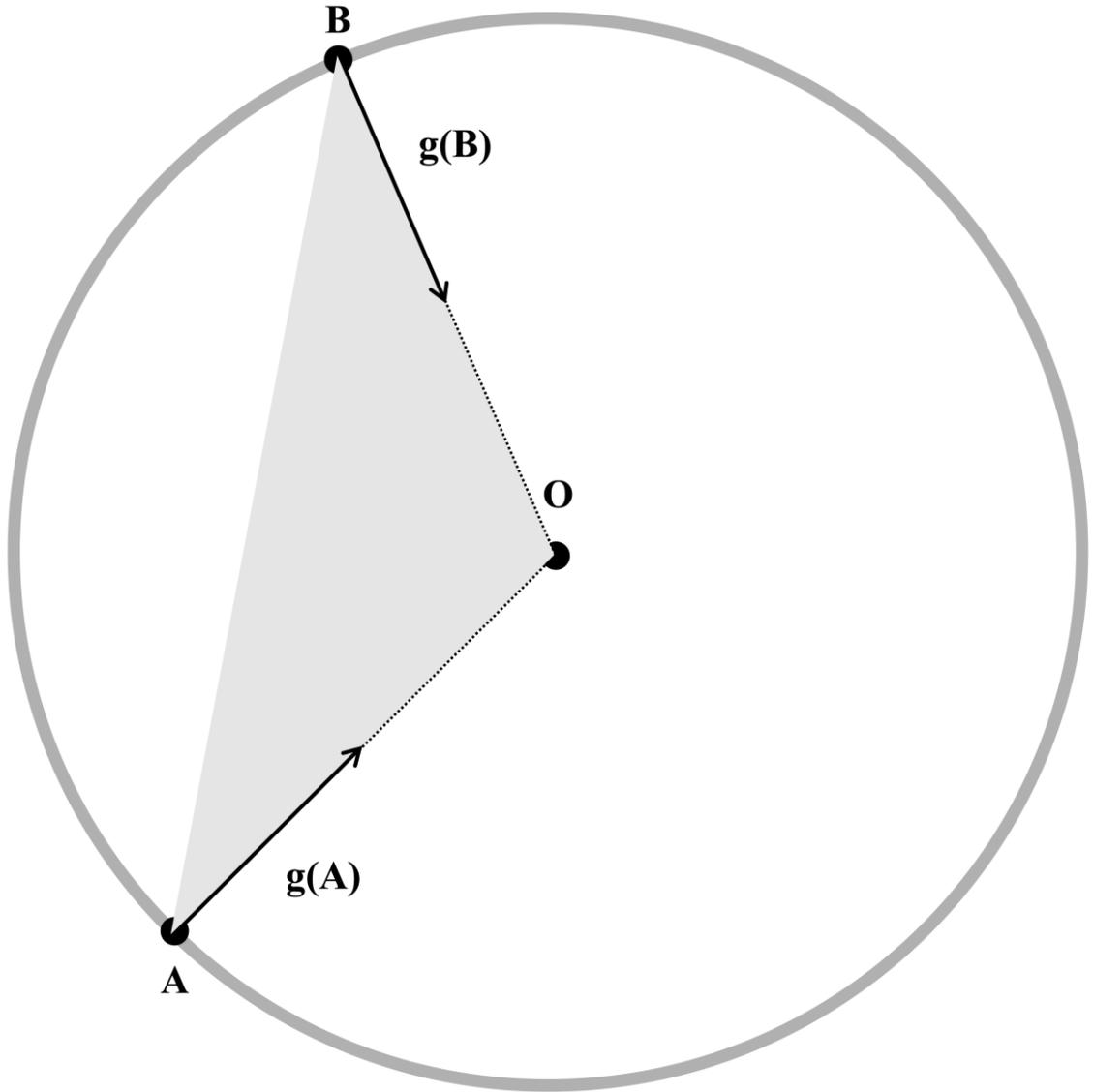

Figure 1

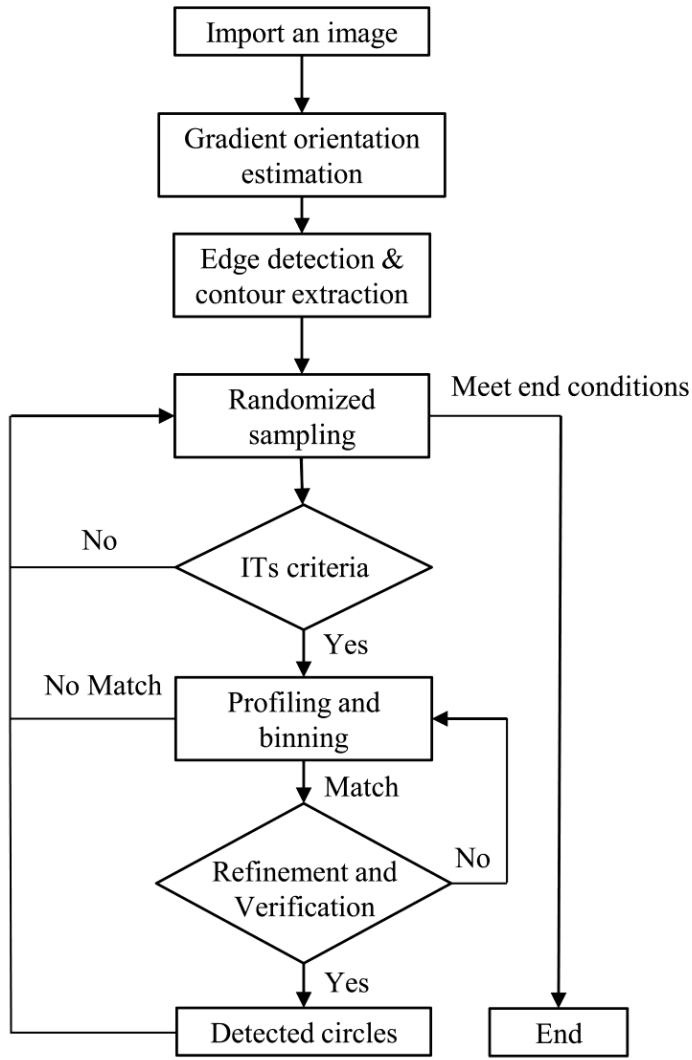

Figure 2

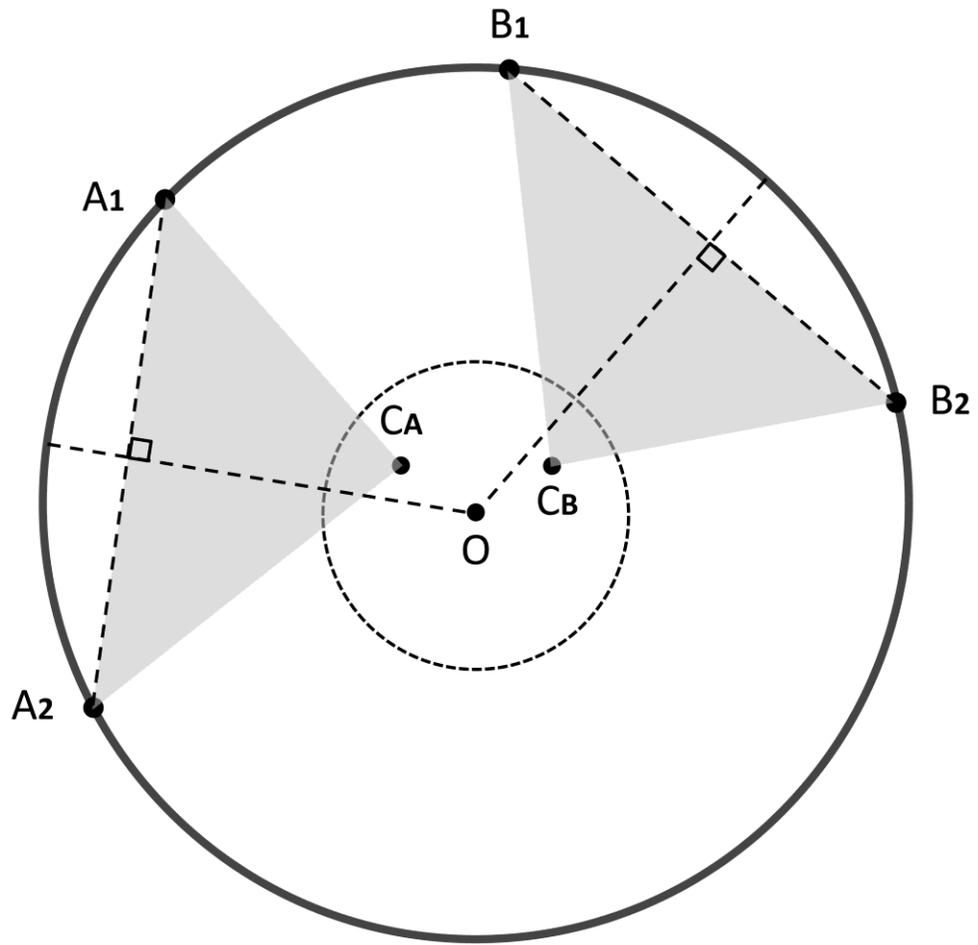

Figure 3

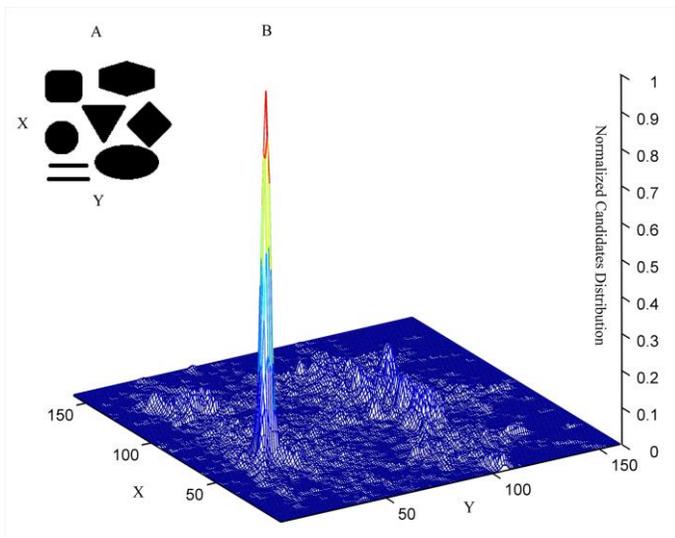

Figure 4

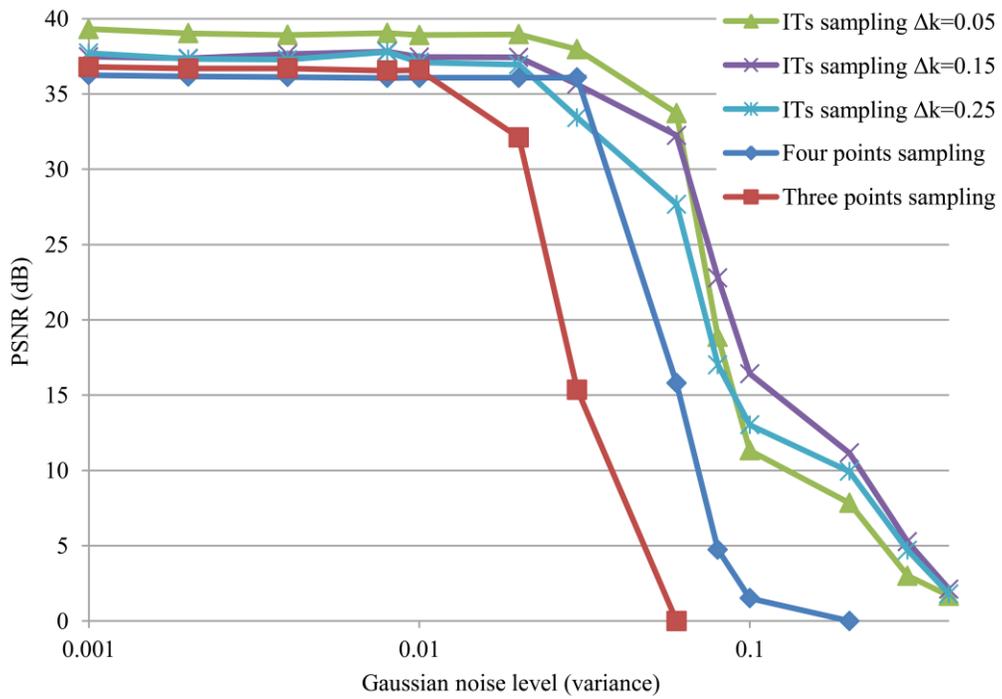

Figure 5

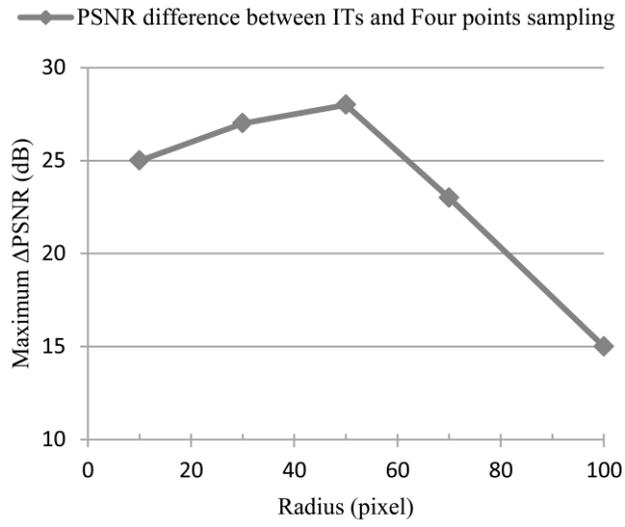

Figure 6

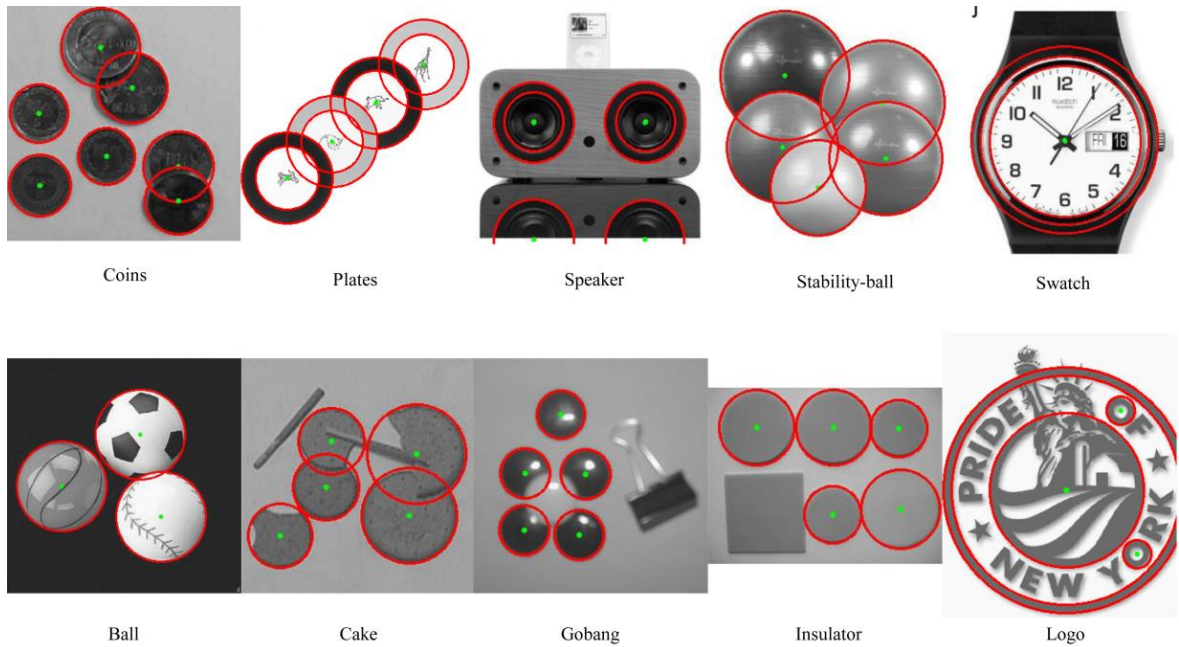

Figure 7

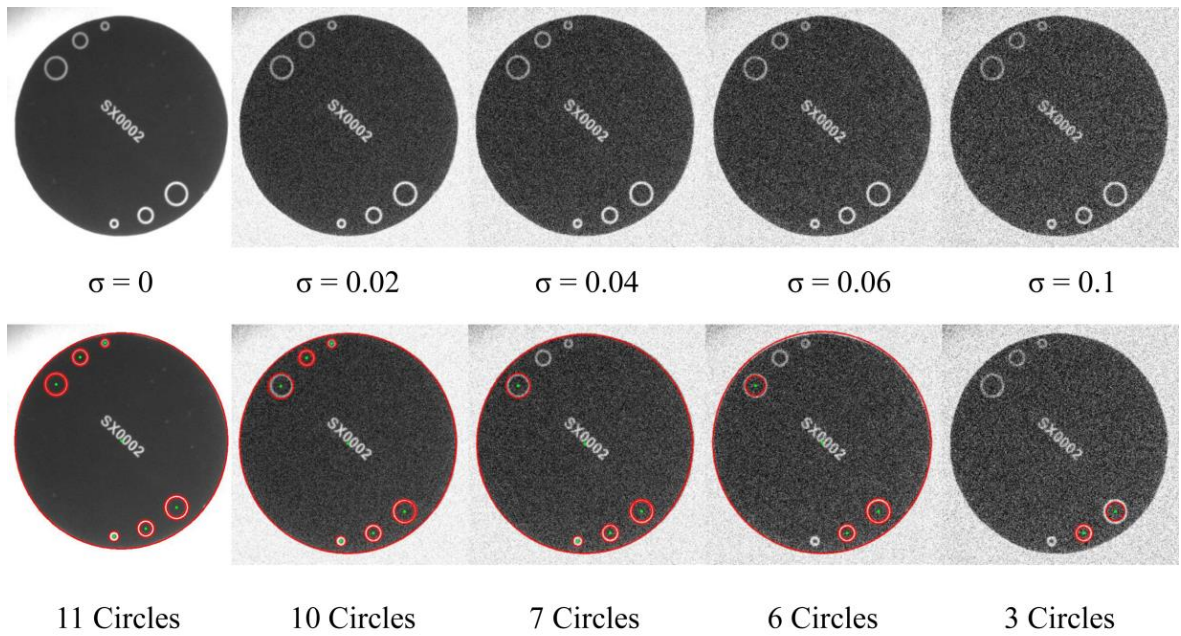

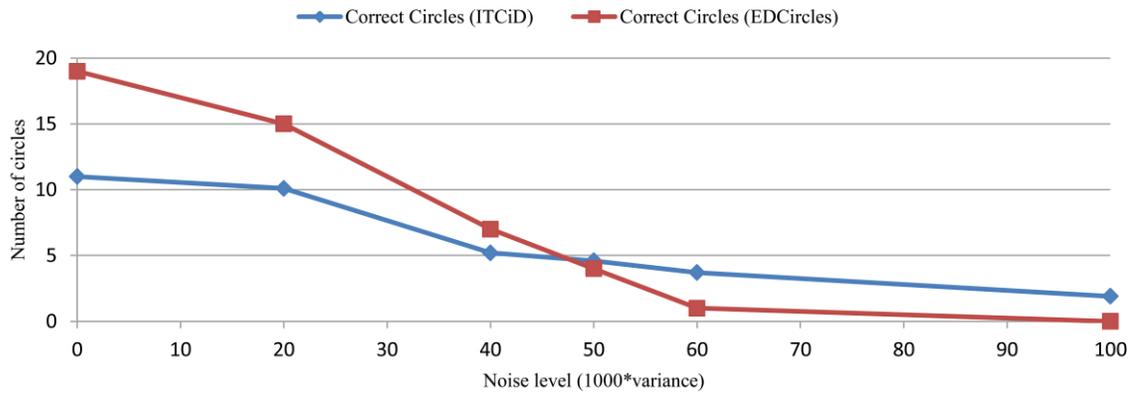

Figure 8

Figure 9

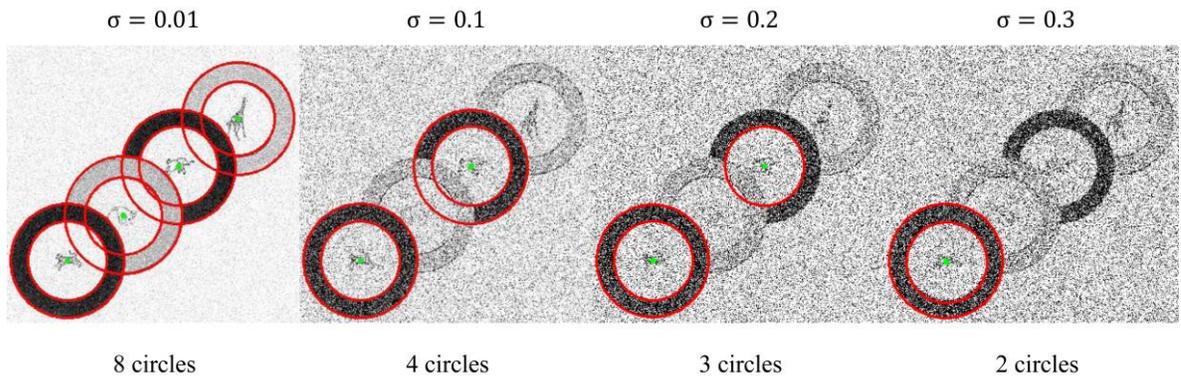

$\sigma = 0.01$     $\sigma = 0.1$     $\sigma = 0.2$     $\sigma = 0.3$

8 circles          4 circles          3 circles          2 circles

Figure 10

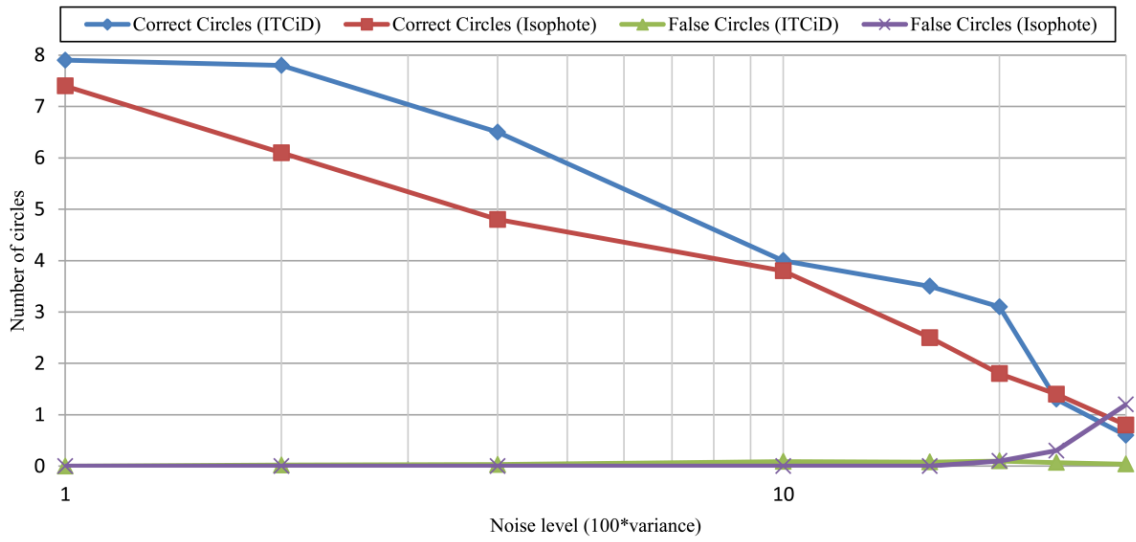

Figure 11